\documentclass{elsart}
\usepackage{algorithmic}
\usepackage{algorithm}
\usepackage{graphicx,amssymb}
\usepackage{enumerate}
\usepackage[switch,pagewise]{lineno}
\usepackage{url}
\usepackage{cite}
\usepackage{amsmath}
\usepackage[usenames]{color}
\usepackage{subfig}
\usepackage{multirow} 
\usepackage{booktabs}

\begin{document}

\begin{frontmatter}

\title{Reinforcement learning based local search for grouping problems: A case study on graph coloring}

\author[Angers]{Yangming Zhou}, 
\ead{zhou.yangming@yahoo.com}
\author[Angers,IUF]{Jin-Kao Hao\corauthref{cor}},
\corauth[cor]{Corresponding author.}
\ead{hao@info.univ-angers.fr}
\author[Angers]{B{\'e}atrice Duval}
\ead{bd@info.univ-angers.fr}
\address[Angers]{LERIA, Universit{\'e} d'Angers, 2 Bd Lavoisier, 49045 Angers, France}
\address[IUF]{Institut Universitaire de France, Paris, France}

\maketitle

\begin{abstract}
Grouping problems aim to partition a set of items into multiple mutually disjoint subsets according to some specific criterion and constraints. Grouping problems cover a large class of important combinatorial optimization problems that are generally computationally difficult. In this paper, we propose a general solution approach for grouping problems, i.e., reinforcement learning based local search (RLS), which combines reinforcement learning techniques with descent-based local search. The viability of the proposed approach is verified on a well-known representative grouping problem (graph coloring) where a very simple descent-based coloring algorithm is applied. Experimental studies on popular DIMACS and COLOR02 benchmark graphs indicate that RLS achieves competitive performances compared to a number of well-known coloring algorithms.


\end{abstract}


\end{frontmatter}


\section{Introduction}
\label{sec:Introduction}

Grouping problems aim to partition a set of items into a collection of mutually disjoint subsets according to some specific criterion and constraints. Grouping problems naturally arise in numerous domains. Well-known grouping problems include, for instance, graph coloring (GCP) \cite{GareyJohnson1979,Galinier1999,Lewis2009,Elhag2O15}, timetabling \cite{Lewis2007,Elhag2O15}, bin packing \cite{Falkenauer1998,Quiroz-Castellanos2015}, scheduling \cite{Kashan2013} and clustering \cite{Agust2012}. Formally, given a set $V$ of $n$ distinct items, the task of a grouping problem is to partition the items of set $V$ into $k$ different groups $g_i$ $(i=1,\ldots,k)$ ($k$ can be fixed or variable), such that $\cup^k_{i=1} g_i = V$ and $g_i \cap g_j = \emptyset, i \neq j$ while taking into account some specific constraints and optimization objective. For instance, the graph coloring problem is to partition the vertices of a given graph into a minimum number of $k$ color classes such that adjacent vertices must be put into different color classes.


According to whether the number of groups $k$ is fixed in advance, grouping problems can be divided into constant grouping problems or variable grouping problems \cite{Kashan2013}. In some contexts, the number of groups $k$ is a fixed value of the problem, such as identical or non-identical parallel-machines scheduling problem, while in other settings, $k$ is variable and the goal is to find a feasible grouping with a minimum number of groups, such as the bin packing problem and graph coloring problem. Grouping problems can also be classified according to the types of the groups. A grouping problem with identical groups means that all groups have similar characteristics, thus naming of the groups is irrelevant. Aforementioned examples such as identical parallel-machines scheduling, bin-packing and graph coloring belong to this category. Another category of grouping problems have non-identical groups where the groups are of different characteristics. Hence, swapping items between two groups will result in a new grouping, such as the non-identical parallel-machines scheduling problem.


Many grouping problems, including the examples mentioned above are NP-hard, thus computationally challenging. Due to the high computational complexity of these problems, exponential times are expected for any algorithm to solve such a problem exactly. On the other hand, heuristic and meta-heuristic methods are often employed to find  satisfactory sub-optimal solutions in acceptable computing time, but without provable optimal guarantee of the attained solutions. A number of heuristic approaches for grouping problems, in particular based on genetic algorithms, have been proposed in the literature with varying degrees of success \cite{Falkenauer1998,Galinier1999,Quiroz-Castellanos2015}. These approaches are rather complex since they are population-based and often hybridized with other search methods like local optimization.


In this work, we are interested in investigating a general purpose local search methodology for grouping problems which employs machine learning techniques to process information collected from the search process with the purpose of improving the performance of heuristic algorithms. Indeed, previous work has demonstrated that machine learning can contribute to improve optimization methods \cite{Baluja2000,Battiti2014,Hafiz2016}. Existing research in these areas has pursued different objectives.

\begin{itemize}

\item Algorithm selection and analysis. For instance, Hutter et al. used machine learning techniques such as random forests and approximate Gaussian process to model algorithm's runtime as a function of problem-specific instance features. This model can predict algorithm runtime for the propositional satisfiability problem, travelling salesperson  problem and mixed integer programming problem \cite{Hutter2014}.

\item Learning generative models of solutions. For example, Ceberio et al. introduced the Plackett-Luce probability model to the framework of estimation of distribution algorithms and applied it to solve the linear order problem and the flow-shop scheduling problem \cite{Ceberio2013}. 

\item Learning evaluation functions. For instance, Boyan and Moore proposed the STAGE algorithm to learn an evaluation function which predicts the outcome of a local search algorithm such as hill-climbing or Walk-SAT, as a function of state features along its search trajectories. The learned evaluation function is used to bias future search trajectories towards better solutions \cite{Boyan2001}.

\item Understanding the search space. For example, Porumbel et al. used multidimensional scaling techniques to explore the spatial distribution of the local optimal solutions visited by tabu search, thus improving local search algorithms for the graph coloring problem \cite{Porumbel2010a}.For the same problem, the authors of \cite{HamiezHao1993} used the results of an analysis of legal $k$-colorings to help finding solutions with fewer colors. 
\end{itemize}

In this paper, we present the reinforcement learning based local search (RLS) approach for grouping problems, which combines reinforcement learning techniques with a descent-based local search procedure. Our proposed RLS approach belongs to the above-mentioned category of learning generative models of solutions. For a grouping problem with its $k$ groups, we associate to an item a probability vector with respect to each possible group and determine the group of the item according to the probability vector. Once all items are assigned to their groups, a grouping solution is generated. Then, the descent-based local search procedure is invoked to improve this solution until a local optimum is attained. Afterward, the probability vector of each item is updated by comparing the group of the item in the starting solution and in the attained local optimum solution. If an item stays in its original group, then we reward the selected group of the item, otherwise we penalize the original group and compensate the new group (i.e., expected group). There are two key issues that need to be considered, i.e., how do we select a suitable group for each item according to the probability vector, and how do we smooth the probabilities to avoid potential search traps. To handle these issues, we design two strategies: a hybrid group selection strategy that uses a noise probability to switch between random selection and greedy selection; and a probability smoothing mechanism able to forget old decisions.

To evaluate the viability of the proposed RLS method, we use the well-known graph coloring problem (GCP) as a case study. GCP is one representative grouping problem which has been object of intensive studies in the past decades. We show computational experiments on both DIMACS and COLOR02 benchmark graphs. Computational results demonstrate that the proposed approach, despite its simplicity, achieves competitive performances on most tested instances compared to many existing algorithms. With an analysis of three important issues of RLS, we show the effectiveness of combining reinforcement learning and descent-based local search. We also assess the contribution of the probability smoothing technique to the performance of RLS. 

The rest of the paper is organized as follows. Section \ref{sec:RLHS} provides an introduction of reinforcement learning and its applications to enhance heuristic search. The proposed RLS method is described in Section \ref{sec:RLS}. Section \ref{sec:Experiments} is dedicated to computational assessments and comparisons of RLS applied to the graph coloring problem. Concluding comments and future research directions are discussed in Section \ref{sec:Conclusion}.

\section{Reinforcement learning and heuristic search}
\label{sec:RLHS}

In this section, we briefly introduce the principles of reinforcement learning (RL) and provide a review of some representative examples of using reinforcement learning to solve combinatorial optimization problems.

\subsection{Reinforcement learning}
\label{subsec:RL}

Reinforcement learning is a learning pattern, which aims to learn optimal actions from a finite set of available actions through continuously interacting with an unknown environment. In contrast to supervised learning techniques, reinforcement learning does not need an experienced agent to show the correct way, but adjusts its future actions based on the obtained feedback signal from the environment \cite{Gosavi2009}.

There are three key elements in a RL agent, i.e., states, actions and rewards. At each instant a RL agent observes the current state, and takes an action from the set of its available actions for the current state. Once an action is performed, the RL agent changes to a new state, based on transition probabilities. Correspondingly, a feedback signal is returned to the RL agent to inform it about the quality of its performed action. 

\subsection{Reinforcement learning and heuristic search}
\label{subsec:RRLHS}

There are a number of studies in the literature where reinforcement learning techniques are put at the service of heuristic algorithms for solving combinatorial problems. Reinforcement learning techniques in these studies have been explored at three different levels.

\textit{Heuristic level} where RL is directly used as a heuristic to solve optimization problems. In this case, RL techniques are used to learn and directly assign values to the variables. For example, the authors of \cite{Miagkikh1999} proposed to solve combinatorial optimization problems based on a population of RL agents. Pairs of variable and value are considered as the RL states, and the branching strategies as the actions. Each RL agent is assigned a specific area of the search space where it has to learn and find good local solutions.

\textit{Meta-heuristic level} where RL is integrated into a meta-heuristic. There are two types of these algorithms. Firstly, RL is used to learn properties of good initial solutions or an evaluation function that guides a meta-heuristic toward high quality solutions. For example, RL is employed to learn a new evaluation function over multiple search trajectories of the same problem instance and alternates between using the learned and the original evaluation function \cite{Boyan2001}. Secondly, RL learns the best neighborhoods or heuristics to build or change a solution during the search, so that a good solution can be obtained at the end. For instance, Xu et al. \cite{Xu2009} proposed a formulation of constraint satisfaction problems as a RL task. A number of different variable ordering heuristics are available, and RL learns which one to use, and when to use it. 

\textit{Hyper-heuristic level} where RL is used as a component of a hyper-heuristic. Specifically, RL is integrated into selection mechanisms and acceptance mechanisms in order to select a suitable low-level heuristic and determine when to accept a move respectively. For example, Burke et al. \cite{Burke2003} presented a hyper-heuristic in which the selection of low-level heuristics makes use of basic reinforcement learning principles combined with a tabu search mechanism. The algorithm increases or decreases the rank of the low-level heuristics when the objective function value is improving or deteriorating. Two other examples can be found in \cite{Guoetal2013,Sghiretal2015} where RL is used to schedule several search operators (crossovers, local search...) under the genetic and multi-agent based optimization frameworks. 

Both meta-heuristic level and hyper-heuristic level approaches attempt to replace the random component of an algorithm with a RL component to obtain an informed decision mechanism. Based on the above-classification, our proposed RLS approach belongs to first type of the meta-heuristic level category. Specifically, RLS combines reinforcement learning techniques with descent-based local search with the purpose of learning properties of good initial solutions.

\section{Reinforcement learning based local search for grouping problems}
\label{sec:RLS}

Grouping problems aim to partition a set of items into $k$ disjoint groups according to some imperative constraints and an optimization criterion. For our RLS approach, we suppose that the number of groups $k$ is given in advance. Note that such a assumption is not necessarily restrictive. In fact, to handle a grouping problem with variable $k$, one can repetitively run RLS with different $k$ values. We will illustrate this approach on the graph coloring problem in Section \ref{sec:Experiments}.


\subsection{Main scheme}
\label{subsec:mainscheme}

By combining reinforcement learning techniques with a solution improvement procedure, our proposed RLS approach is composed of four keys components: a descent-based local search procedure, a group selection strategy, a probability updating mechanism (i.e., reinforcement learning mechanism), and a probability smoothing technique. 

\begin{figure}[!htbp]
\centering
\includegraphics[width=0.5\textwidth]{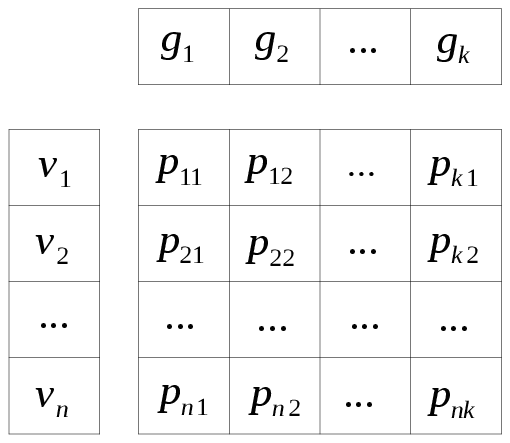}
\caption{Probability matrix $P$}\label{fig:probabilityMatrix}
\end{figure}

We define a probability matrix $P$ of size $n \times k$ ($n$ is the number of items and $k$ is the number of groups, see Figure \ref{fig:probabilityMatrix} for an example). An element $p_{ij}$ denotes the probability that the $i$-th item $v_i$ selects the $j$-th group $g_j$ as its group. Therefore, the $i$-th row of the probability matrix defines the probability vector of the $i$-th item and is denoted by $p_i$. At the beginning, all the probability values in the probability matrix are set as $1/k$. It means that all items select a group from the available $k$ groups with equal probability.

At instant $t$, each item $v_i$, $i \in \{1, 2, ..., n\}$ selects one suitable group $g_j$, $j \in \{1, 2, ..., k\}$ by applying a group selection strategy (Section \ref{subsec:groupselect}) based on its probability vector $p_i(t)$. Once all the items are assigned to their groups, a grouping solution $S_t$ is obtained. Then, this solution is improved by a descent-based local search procedure to attain a local optimum denoted by $\hat{S_t}$ (Section \ref{subsec:hillclimb}). By comparing the solution $S_t$ and the improved solution $\hat{S_t}$, we update the probability vector of each item based on the following rules (Section \ref{subsec:probabilityupdating}):
\begin{enumerate}[(a)]
\item If the item stays in its original group, then we reward the selected group.
\item If the item is moved to a new group, then we penalize the selected group and compensate its new group (i.e., expected group).
\end{enumerate}

\begin{figure}[!htbp]
\centering
\includegraphics[width=1.0\textwidth]{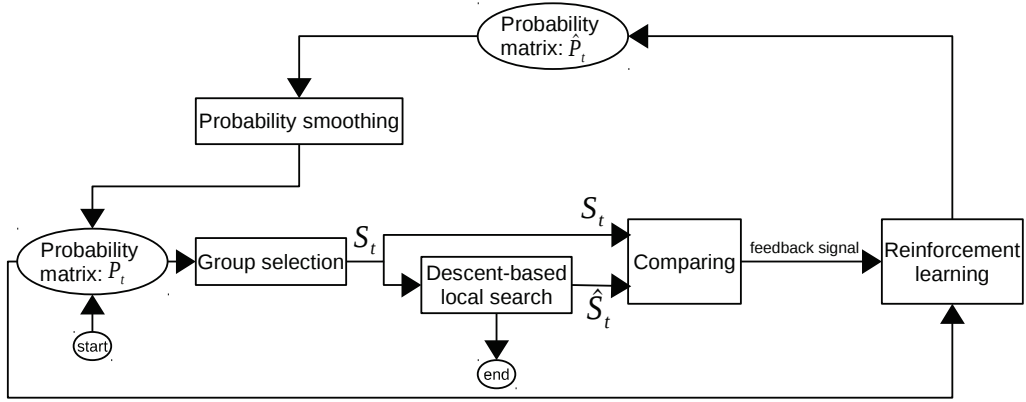}
\caption{A schematic diagram of RLS for grouping problems. From a starting solution generated according to the probability matrix, RLS iteratively runs until its meets its stop condition (see Sections \ref{subsec:groupselect}-\ref{subsec:probabilitysmooth} for more details)}\label{fig:diagram}
\end{figure}

Next, we apply a probability smoothing technique to smooth each item's probability vector (Section \ref{subsec:probabilitysmooth}). Hereafter, RLS iteratively runs until a predefined stop condition is reached (e.g., a legal solution is found or the number of iterations without improvement exceeds a maximum allowable value). The schematic diagram of RLS for grouping problems is depicted in Figure \ref{fig:diagram} while its algorithmic pseudo-code is provided in Algorithm \ref{algorithm:RLS}. In the following subsections, the four key components of our RLS approach are presented in detail.

\begin{algorithm}
\begin{small}
 \caption{Pseudo-code of our RLS for grouping problems.}
 \label{algorithm:RLS}
 \begin{algorithmic}[1]
 	\STATE \sf \textbf{Input}:\\
 		\ \ \ $G$: a grouping problem instance;\\
 		\ \ \ $k$: the number of available groups;\\
 	\STATE \textbf{Output}: the best solution $S^{*}$ found so far;
	\FORALL {$v_i , i = 1, 2, ..., n$}
		\STATE $P_0 =[p_{ij} = 1/k]_{j = 1, 2, ..., k}$;
	\ENDFOR
	\REPEAT
		\STATE $S_t \leftarrow groupSelecting(P_{t-1}, \omega)$; \ \ \ \ \ \ \ \ \ \ \ \ \ \ \ \ \ \ \ \ \ \ \ \ \ \ \ \ \ \ \ \ \ \ \ \ \  /$*$ Section \ref{subsec:groupselect} $*$/

		\STATE $\hat {S}_t \leftarrow DB-LS(S_t)$; \ \ \ \ \ \ \ \ \ \ \ \ \ \ \ \ \ \ \ \ \ \ \ \ \ \ \ \ \ \ \ \ \ \ \ \ \ \ \ \ \ \ \ \ \ \ \ \ \ \ \ /$*$ Section \ref{subsec:hillclimb} $*$/
		\STATE $P_t \leftarrow probabilityUpdating(P_{t-1}, S_t, \hat{S}_t , \alpha, \beta, \gamma)$; \ \ \ \ \ \ \ \ \ \ \ \ \ \ \ \ /$*$ Section \ref{subsec:probabilityupdating} $*$/
		\STATE $P_t \leftarrow probabilitySmoothing(P_t , p_0 , \rho)$; \ \ \ \ \ \ \ \ \ \ \ \ \ \ \ \ \ \ \ \ \ \ \ \ \ \ \  /$*$ Section \ref{subsec:probabilitysmooth} $*$/
	\UNTIL{Stop condition met}
\end{algorithmic}
\end{small}
\end{algorithm}

\subsection{Group selection}
\label{subsec:groupselect}

At each iteration of RLS, each item $v_i$ needs to select a group $g_j$ from the $k$ available groups according to its probability vector $p_i$. We consider four possible group selection strategies:
\begin{itemize}
	\item Random selection: the item selects its group at random (regardless of its probability vector). As this selection strategy does not use any useful information collected from the search history, it is expected that this strategy would not perform well. 
	\item Greedy selection: the item always selects the group $g_j$ such that the associated probability $p_{ij}$ has the maximum value. This strategy is intuitively reasonable, but may cause the algorithm to be trapped rapidly. 
	\item Roulette wheel selection: the item selects its group based on its probability vector and the chance for the item to select group $g_j$ is proportional to the probability $p_{ij}$. Thus a group with a large (small) probability has more (less) chance to be selected.
	\item Hybrid selection: this strategy combines the random selection and greedy selection strategies in a probabilistic way; with a noise probability $\omega$, random selection is applied; with probability $1 - \omega$, greedy selection is applied.
\end{itemize} 


As we show in Section \ref{Comparison of different group selection strategies}, the group selection strategy greatly affects the performance of the RLS approach. After experimenting the above strategies, we adopted the hybrid selection strategy which combines randomness and greediness which are controlled by the noise probability $\omega$.  The purpose of selecting a group with maximum probability (greedy selection) is to make an attempt to correctly select the group for an item that is most often falsified at a local optimum. Selecting such a group for this item may help the search to escape from the current trap. On the other hand, using the noise probability has the advantage of flexibility by switching back and forth between greediness and randomness. Also, this allows the algorithm to occasionally move away from being too greedy. This hybrid group selection strategy proves to be better than the roulette wheel selection strategy, as confirmed by the experiments of Section \ref{Comparison of different group selection strategies}.

\subsection{Descent-based local search for solution improvement}
\label{subsec:hillclimb}

Even if any optimization procedure can be used to improve a given starting grouping solution. For the reason of simplicity, we employ a simple and fast descent-based local search (DB-LS) procedure in this work. To explore the search space, DB-LS iteratively makes transitions from the incumbent solution to a neighboring solution according to a given neighborhood relation such that each transition leads to a better solution. This iterative improvement process continues until no improved solution exists in the neighborhood in which case the incumbent solution corresponds to a local optimum with respect to the neighborhood. 

Let $\Omega$ denote the search space of the given grouping problem. Let $N: \Omega \rightarrow 2^\Omega$ be the neighborhood relation which associates to each solution $S \in \Omega$ a subset of solutions $N(S) \subset \Omega$ (i.e., $N(S)$ is the set of neighboring solutions of $S$). Typically, given a solution $S$, a neighboring solution can be obtained by moving an item of $S$ from its current group to another group. Let $f: \Omega \rightarrow \mathbb{R}$ be the evaluation (or cost) function which measures the quality or cost of each grouping solution. The pseudo code of Algorithm \ref{algorithm:hillClimbing} displays the general DB-LS procedure.


\begin{algorithm}
\begin{small}
 \caption{Pseudo-code of descent-based local search procedure}
 \label{algorithm:hillClimbing}
 \begin{algorithmic}[1]
  	\STATE \sf \textbf{Input}: $S$ - an initial candidate grouping solution;\\
 	\STATE \textbf{Output}: $S^{*}$ - the local optimum solution attained;
 	\STATE $f(S^*) = f(S)$;
 	\REPEAT
		\STATE choose a best neighbor $S^{''}$ of $S$ such that
			\STATE $S^{''} = \arg \min_{S^{'} \in N(S)} f(S)$;
		\STATE $S^* = S^{''}$;
		\STATE $f(S^*) = f(S^{''})$
		\STATE $S = S^*$;
	\UNTIL {$f(S^{''}) \geqslant f(S^*)$}
 \end{algorithmic}
\end{small}
\end{algorithm}

Descent-based local search can find a local optimum quickly. However, the local optimal solution discovered is generally of poor quality. It is fully possible to improve the performance of RLS by replacing the descent-based local search with a more powerful improvement algorithm. In RLS, we make the assumption that, if the item stays in its original group after the descent-based local search, then the item has selected the right group in the original solution, otherwise its new group in the improved solution would be the right group. This assumption can be considered to be reasonable because the descent-based local search procedure is driven by its cost function and each transition from the current solution to a new (neighboring) solution is performed only when the transition leads to an improvement.

\subsection{Reinforcement learning - probability updating}
\label{subsec:probabilityupdating}

Reinforcement learning is defined as how an agent should take actions in an environment so to maximize some notion of cumulative reward. Reinforcement learning acts optimally through trial-and-error interactions with an unknown environment. Actions may affect not only the immediate reward but also the next situation and all subsequent rewards. The intuition underlying reinforcement learning is that actions that lead to large rewards should be made more likely to recur. In RLS, the problem of selecting the most appropriate group for each item is viewed as a reinforcement learning problem. Through the interactions with the unknown environment, RLS evolves and gradually finds the optimal or a suboptimal solution of the problem.

At instant $t$, we firstly generate a grouping solution $S_t$ based on the current probability matrix $P_t$ (see Section \ref{subsec:mainscheme}). In other words, each item selects one suitable group from the $k$ available groups based on its probability vector (with the group selection strategy of Sect. \ref{subsec:groupselect}). Then solution $S_t$ is improved by the descent-based local search procedure, leading to an improved solution $\hat{S_t}$. Now, for each item $v_i$, we compare its groups in $S_t$ and  $\hat{S_t}$. If the item stays in its original group (say $g_u$), we reward the selected group $g_u$ (called correct group) and update its probability vector $p_i$ according to Eq. (\ref{equ:reward}):
\begin{equation} \label{equ:reward}
p_{ij}(t+1) = 
\begin{cases}
{\alpha + (1-\alpha)p_{ij}(t)} & j = u \\
{(1-\alpha)p_{ij}(t)} & \text{otherwise}.
\end{cases}
\end{equation}
where $\alpha~(0 < \alpha < 1)$ is a reward factor. When item $v_i$ moves from its original group $g_u$ of solution $S_t$ to a new group (say $g_v, v\neq u$) of the improved solution $\hat{S_t}$, we penalize the discarded group $g_u$ (called incorrect group),  compensate the new group $g_v$ (called expected group) and finally update its probability vector $p_i$ according to Eq. (\ref{equ:penalizeAndCompensate}):
\begin{equation} \label{equ:penalizeAndCompensate}
p_{ij}(t+1) = 
\begin{cases}
(1-\gamma)(1-\beta)p_{ij}(t) & j = u \\
\gamma + (1-\gamma){\frac{\beta}{k-1}} + (1-\gamma)(1-\beta)p_{ij}(t) & j = v \\
(1-\gamma){\frac{\beta}{k-1}} + (1-\gamma)(1-\beta)p_{ij}(t) & \text{otherwise}.
\end{cases}
\end{equation} 
where $\beta~(0 < \beta < 1)$ and $\gamma~(0 < \gamma < 1)$ are a penalization factor and compensation factor respectively. This process is repeated until each item can select its group correctly. The update of the complete probability matrix $P$ is bounded by $\mathcal{O}(n \times k)$ in terms of time complexity.

It is necessary to note that our learning scheme is different from general reinforcement learning schemes such as linear reward-penalty, linear reward-inaction and linear reward-$\epsilon$-penalty. The philosophy of these schemes is to increase the probability of selecting an action in the event of success and decrease it when receives a failed signal. Unlike these general schemes, our learning scheme not only rewards the correct group and penalizes the incorrect group, but also compensates the expected group.

\subsection{Reinforcement learning - probability smoothing}
\label{subsec:probabilitysmooth}

The intuition behind the probability smoothing technique is that old decisions that were made long ago are no longer helpful and may mislead the current search. Therefore, these aged decisions should be considered less important than the recent ones. In addition, all items are required to correctly select their suitable groups in order to produce a legal grouping solution. It is not enough that only a part of items can correctly select their groups. Based on these two reasons, we introduce a probability smoothing technique to reduce the group probabilities periodically.

Our probability smoothing strategy is inspired by forgetting mechanisms in smoothing techniques in clause weighting local search algorithms for satisfiability (SAT) \cite{Hutter2002,Ishtaiwi2005}. Based on the way that weights are smoothed or forgotten, there are four available forgetting or smoothing techniques for MVC and SAT: 
\begin{itemize}
\item Decrease one from all clause weights which are greater than one such as PAWS \cite{Thornton2004}.

\item Pull all clause weights to their mean value using the formula $w_i = \rho \cdot w_i + (1-\rho)\cdot \overline{w_i}$ like ESG \cite{Schuurams2001} and SAPS \cite{Hutter2002}.

\item Transfer weights from neighboring satisfied clauses to unsatisfied ones like DDWF \cite{Ishtaiwi2005}.

\item Reduce all edge weights using the formula $w_i = \lfloor \rho \cdot w_i \rfloor$ when the average weight achieves a threshold like NuMVC \cite{Cai2013}.
\end{itemize}

The probability smoothing strategy adopted in our RLS approach works as follows (see Algorithm \ref{algorithm:probabilitySmooth}). For an item, each possible group is associated with a value between $0$ and $1$ as its probability, and each group probability is initialized as $1/k$. At each iteration, we adjust the probability vector based on the obtained feedback information (i.e., reward,  penalize or compensate a group). Once the probability of a group in a probability vector achieves a given threshold (i.e., $p_0$), it is reduced by multiplying a smoothing coefficient (i.e., $\rho<1$) to forget some earlier decisions. It is obvious that the smoothing technique used in RLS is different from the above-mentioned four techniques. To the best of our knowledge, this is the first time a smoothing technique is introduced into local search algorithms for grouping problems.

\begin{algorithm}
\begin{small}
 \caption{Pseudo-code of the probability smoothing procedure}
 \label{algorithm:probabilitySmooth}
 \begin{algorithmic}[1]
 	\STATE \sf \textbf{Input}:\\
 		\	\	\ $P_t$: probability matrix at instant $t$;\\
 		\	\	\ $p_0$: smoothing probability;\\
 		\	\	\ $\rho$: smoothing coefficient;\\
 	\STATE \textbf{Output}: new probability matrix $P_t$ after smoothing;
	 	\FOR {$i = 1$ to $ n$}
	\STATE $p_{iw} = max\{p_{ij}, j = 1,2,...,k\}$;
		\IF{$p_{iw} > p_0$}
	 	\FOR {$j = 1$ to $ k$}
				\IF{$j = w$}
					\STATE $p_{ij}(t)=\rho \cdot p_{ij}(t-1)$;
				\ELSE
					\STATE $p_{ij}(t)= \frac{1-\rho}{k-1} \cdot p_{iw}(t-1)+p_{ij}(t-1)$;
				\ENDIF
			\ENDFOR
		\ENDIF
	\ENDFOR
 \end{algorithmic}
\end{small}
\end{algorithm}

\section{RLS applied to graph coloring: a case study}
\label{sec:Experiments}

This section presents an application of the proposed RLS method to the well-known graph coloring problem which is a typical grouping problem. After presenting the descent-based local search procedure for the problem, we first conduct an experimental analysis of the RLS approach by investigating the influence of its three important components, i.e., the reinforcement learning mechanism, the probability smoothing technique and the group selection strategy. Then we present computational results attained by the proposed RLS method in comparison with a number of existing local search algorithms over well-known DIMACS and COLOR02 benchmark instances.

\subsection{Graph coloring and local search coloring algorithm}
\label{Graph coloring problem}

GCP is one of the most studied combinatorial optimization problems \cite{GareyJohnson1979}. GCP  is also a nice representative of grouping problems. Given an undirected graph $G = (V, E)$, where $V$ is the set of $\vert V \vert = n$ vertices and $E$ is the set of $\vert E \vert = m$ edges, a legal $k$-coloring of $G$ is a partition of $V$ into $k$ mutually disjoint groups or color classes such that two vertices linked by an edge must belong to two different color classes. GCP is to determine the smallest $k$ for a graph $G$ such that a legal $k$-coloring exists. This minimum number of groups (i.e., colors) required for a legal coloring is the \textit{chromatic number} $\chi(G)$. When the number of color classes $k$ is fixed, the problem is called $k$-coloring problem ($k$-GCP for short). As a grouping problem, items correspond to vertices and groups correspond to color classes.

Notice that GCP can be approximated by solving a series of $k$-GCP (with decreasing $k$) as follows \cite{Galinier2013}. For a given $G$ and a given $k$, we use our RLS approach to solve $k$-GCP by seeking a legal $k$-coloring. If such a coloring is successfully found, we decrease $k$ and solve the new $k$-GCP again. We repeat this process until no legal $k$-coloring can be reached. In this case, the last $k$ for which a legal $k$-coloring has been found represents an approximation (upper bound) of the chromatic number of $G$. This general solution approach has been used in many coloring algorithms including most of those reviewed below, and is adopted in our work. 



Given the theoretical and practical interest of GCP, a huge number of coloring algorithms have been proposed in the past decades \cite{Galinier2013,JohnsonTrick1996,MalagutiToth2009,AkbariTorkestani2011}. Among them, algorithms based on local search are certainly the most popular approaches, like simulated annealing (SA) \cite{Johnson1991}, tabu search (TS) \cite{Hertz1987, Galinier1999}, guided local search (GLS) \cite{Chiarandini2005}, iterated local search (ILS) \cite{Chiarandini2002}, quantum annealing algorithms \cite{Titiloye2011} and focused walk based local search (FWLS) \cite{Wu2013}. Population-based hybrid algorithms represent another class of complex approaches which typically combine local search and dedicated recombination crossover operators \cite{Fleurent1996,Galinier1999,Lu2010,Malagutietal2008,Porumbel2010b}. Recent surveys of algorithms for GCP can be found in \cite{Galinier2013,MalagutiToth2009}.


To apply the proposed RLS approach to $k$-GCP, we need to specify three important ingredients of the descent-based local search in RLS, i.e., the search space, the neighborhood and the evaluation function. First, a legal or illegal $k$-coloring can be represented by $S = \{g_1 , g_2 , ..., g_k\}$ such that $g_i$ is the group of vertices receiving color $i$. Therefore, the search space $\Omega$ is composed of all possible legal and illegal $k$-colorings. The evaluation function $f(S)$ counts the number of conflicting edges inducted by $S$ such that:
\begin{equation}
	f(S)=\sum_{\{u,v\} \in E} \delta(u,v)
\end{equation}
where $\delta(u,v)=1$, if $u \in g_i, v \in g_j$ and $i=j$, and otherwise $\delta(u,v)=0$. Accordingly, a candidate solution $S$ is a legal $k$-coloring $S$ if $f(S)=0$. 

The neighborhood of a given $k$-coloring is constructed by moving a conflicting vertex $v$ from its original group $g_i$ to another group $g_j (i \neq j)$ \cite{Galinier1999}. Therefore, for a $k$-coloring $S$ with cost $f(S)$, the size of the neighborhood is bounded by $\mathcal{O}(f(S) \times k)$. To evaluate each neighboring solution efficiently, our descent-based local search adopts the fast incremental evaluation technique introduced in \cite{Fleurent1996,Galinier1999}. The principle is to maintain a gain matrix which records the variation $\Delta = f(S') - f(S)$ between the incumbent solution $S$ and every neighboring solution $S'$. After each solution transition from $S$ to $S'$, only the affected elements of the gain matrix are updated accordingly.

The descent-based local search procedure starts then with a random solution taken from the search space $\Omega$ and iteratively improves this solution by a neighboring solution of better quality according to the evaluation function $f$. This process stops either when a legal $k$-coloring is found (i.e., a solution with $f(S)=0$, or no better solution exists among the neighboring solutions (in this later case, a local optimum is reached).

\subsection{Benchmark instances and experimental settings}
\label{Experimental settings}

We show extensive computational results on two sets of the well-known DIMACS\footnote{Publicly available at: ftp://dimacs.rutgers.edu/pub/challenge/graph/benchmarks/color/} and COLOR02\footnote{Publicly available at: http://mat.gsia.cmu.edu/COLOR02/} coloring benchmark instances. These instances are the most widely used benchmark instances for assessing the performance of graph coloring algorithms.

The used DIMACS graphs can be divided into six types: 
\begin{itemize}
\item \textit{Standard random graphs} are denoted as \texttt{DSJC}$n.x$, where $n$ is the number of vertices of the graph. The chromatic number $\chi$ of these graphs are unknown. 

\item \textit{Random geometric graphs} are composed of \texttt{R125.}$x$, \texttt{R250.}$x$, \texttt{DSJR500.}$x$ and \texttt{R1000.}$x$, graphs with letter $c$ being complements of geometric graphs.

\item \textit{Flat graphs} are structured graphs produced based on an equi-partitioning of vertices into $k$ sets. This kind of graphs are denoted as \texttt{flat}$n\_k\_0$, where $n$ and $k$ are the number of vertices and chromatic number respectively.

\item \textit{Leighton graphs} are random graphs of density below 0.25. This kind of graphs are denoted as \texttt{le450}$\_kx$, where 450 is the number of vertices, $k \in \{15, 25\}$ is the chromatic number of the graph, $x \in \{a, b, c, d\}$ is a letter to indicate different graphs with the same characteristics.

\item \textit{Scheduling graphs}, i.e., \texttt{school1} and \texttt{school1\_nsh}.
\item \textit{Latin square graph}, i.e., \texttt{latin\_square\_10}.
\end{itemize}

The used COLOR02 graphs are of three types:
\begin{itemize}
\item \textit{Queen graphs} are highly structured instances and their edge density decreases with their size. The graphs are denoted as \texttt{queen}$x\_x$, where $x \in \{5, 6, 7, 8, 9, 10, 11, 12, 13, 14, 15, 16\}$, with an exception, i.e., \text{queen}$8\_12$.

\item \textit{Mycile graphs} are denoted as \texttt{mycile}$k$, where $k \in \{3, 4, 5, 6, 7\}$. These graphs are based on the Mycielski transformation. 

\item \textit{Miles Graphs} (\texttt{miles}$x$, with $x \in \{250, 500, 750, 1000, 1500\}$) are similar to geometric graphs in that nodes are placed in space with two nodes connected if they are close enough.
\end{itemize}

\begin{table}[!htbp]
\caption{Parameters of Algorithm RLS}
\label{tab:aParameterTable}
\begin{center}
\begin{scriptsize}
\begin{tabular}{lcll}
\toprule[0.75pt]
Parameters 	& Section						  & Description									& Values				\\
\midrule[0.5pt]
$\omega$ 	& \ref{subsec:groupselect}         & noise probability							& 0.200				\\
$\alpha$  	& \ref{subsec:probabilityupdating} & reward factor for correct group 				& 0.100				\\
$\beta$		& \ref{subsec:probabilityupdating} & penalization factor for incorrect group		& (0, 0.45]			\\
$\gamma$		& \ref{subsec:probabilityupdating} & compensation factor for expected group		& 0.300				\\ 
$\rho$ 		& \ref{subsec:probabilitysmooth}   & smoothing coefficient						& 0.500				\\
$p_0$  		& \ref{subsec:probabilitysmooth}   & smoothing threshold							& 0.995				\\
\bottomrule[0.75pt]
\end{tabular}
\end{scriptsize}
\end{center}
\end{table}

Our RLS algorithm was coded in C and compiled using GNU g++ on a machine with an Intel E5-2760 processor (2.8GHz and 2G RAM) under Linux. To obtain our experimental results, each instance was solved 20 times independently with different random seeds. Each execution was terminated when a legal $k$-coloring is found or the number of iterations without improvement reaches its maximum allowable value ($I_{max} = 10^6$). In our experiments, all parameters were fixed except for the penalization factor $\beta$ that varies between 0 and 0.45. Table \ref{tab:aParameterTable} gives the descriptions and settings of the parameters used for our experiments. 



\subsection{Analysis of key components of the RLS approach}
\label{Analysis of key components of the RLS approach}

We first show an analysis of the main ingredients of the RLS approach: reinforcement learning mechanism, probability smoothing technique and group selection strategies. This study allows us to better understand the behavior of the proposed RLS approach and shed lights on its inner functioning.

\subsubsection{Effectiveness of the reinforcement learning mechanism}
\label{Effectiveness of the reinforcement learning mechanism}

To verify the effectiveness of the reinforcement learning mechanism used in RLS, we make a comparison between RLS and its variant $\rm{RLS}_{\rm{0}}$ where we removed the reinforcement learning mechanism from RLS and randomly restart the search when the DB-LS procedure attains a local optimum. 


\begin{table}[!htbp]
\caption{Comparative results of RLS (with reinforcement learning) and $\rm{RLS}_{\rm{0}}$ (without reinforcement learning) on the DIMACS graphs. Smaller $k$ values are better}
\label{tab:aComparisonTableWithHC}
\begin{center}
\begin{scriptsize}
\begin{tabular}{lrrrrrrrrr}
\toprule[0.75pt]
\multicolumn{2}{c}{} & \multicolumn{2}{c}{$\rm{RLS}_{\rm{0}}$} && \multicolumn{2}{c}{RLS} && \multicolumn{1}{c}{Improvement}\\
\cmidrule[0.5pt]{3-4} \cmidrule[0.5pt]{6-7}
Instance 		& $\chi/k^*$ & $k_1$	  & \#hit && $k_2$ 			& \#hit && $k_2 - k_1$ \\
\midrule[0.5pt]
DSJC125.1 		& ?/5   & 6  		  & 12/20 && \textbf{5}		& 20/20 && -1  \\
DSJC125.5 		& ?/17  & 22 		  & 17/20 && \textbf{17}		& 15/20 && -5  \\
DSJC125.9 		& ?/44  & 51 		  & 05/20 && \textbf{44}		& 20/20 && -7  \\
DSJC250.1 		& ?/8   & 11 		  & 20/20 && \textbf{8}		& 20/20 && -3  \\
DSJC250.5 		& ?/28  & 40 		  & 04/20 && \textbf{29}		& 20/20 && -11 \\
DSJC250.9 		& ?/72  & 94 		  & 20/20 && \textbf{75} 	& 01/20 && -19 \\
DSJC500.1 		& ?/12  & 18 		  & 02/20 && \textbf{13} 	& 20/20 && -5  \\
DSJC500.5 		& ?/47  & 74 		  & 04/20 && \textbf{50} 	& 09/20 && -24 \\
DSJC1000.1 		& ?/20  & 32 		  & 18/20 && \textbf{21} 	& 20/20 && -11 \\
DSJR500.1 		& ?/12  & 13 		  & 05/20 && \textbf{12} 	& 20/20 && -1  \\
DSJR500.1c 		& ?/84  & 97 		  & 01/20 && \textbf{85} 	& 02/20 && -12 \\
flat300\_20\_0 	& 20/20 & 44 		  & 02/20 && \textbf{20} 	& 10/20 && -24 \\
flat300\_26\_0 	& 26/26 & 45 		  & 02/20 && \textbf{26} 	& 19/20 && -19 \\
flat300\_28\_0 	& 28/28 & 45 		  & 02/20 && \textbf{32}		& 19/20 && -13 \\
flat1000\_76\_0	& 76/81 & 135		  & 01/20 && \textbf{89} 	& 02/20 && -46 \\
le450\_15a 		& 15/15 & 21 		  & 09/20 && \textbf{15} 	& 19/20 && -6  \\
le450\_15b 		& 15/15 & 21 		  & 20/20 && \textbf{15} 	& 09/20 && -6  \\
le450\_15c 		& 15/15 & 30			  & 01/20 && \textbf{15} 	& 16/20 && -15 \\
le450\_15d 		& 15/15 & 31			  & 19/20 && \textbf{15} 	& 14/20 && -16 \\
le450\_25a 		& 25/25 & 28			  & 18/20 && \textbf{26} 	& 20/20 && -2  \\
le450\_25b 		& 25/25 & 26 		  & 01/20 && \textbf{25} 	& 01/20 && -1  \\
le450\_25c 		& 25/25 & 37 		  & 14/20 && \textbf{26} 	& 13/20 && -11 \\
le450\_25d 		& 25/25 & 36 		  & 03/20 && \textbf{26} 	& 07/20 && -10 \\
R125.1 			& ?/5 	& \textbf{5}  & 20/20 && \textbf{5}  	& 20/20 && 0   \\
R125.1c 			& ?/46 	& \textbf{46} & 15/20 && \textbf{46} 	& 20/20 && 0   \\
R125.5 			& ?/36 	& 42 		  & 07/20 && \textbf{38} 	& 01/20 && -4  \\
R250.1 			& ?/8 	& \textbf{8}  & 20/20 && \textbf{8}  	& 20/20 && 0   \\
R250.1c 			& ?/64 	& 67 		  & 03/20 && \textbf{64} 	& 20/20 && -3  \\
R1000.1 			& ?/20 	& 24 		  & 20/20 && \textbf{21} 	& 20/20 && -3  \\
school1 			& ?/14 	& 39 		  & 05/20 && \textbf{14} 	& 18/20 && -25 \\
school1\_nsh 	& ?/14 	& 36 		  & 02/20 && \textbf{14} 	& 19/20 && -22 \\
latin\_square\_10 & ?/97 & 169 		  & 02/20 && \textbf{99} 	& 10/20 && -70 \\
\bottomrule[0.75pt]
\end{tabular}
\end{scriptsize}
\end{center}
\end{table} 

The investigation was conducted on the 32 DIMACS instances and each algorithm was run 20 times to solve each instance. The comparative results of RLS and $\rm{RLS}_{\rm{0}}$ are provided in Table \ref{tab:aComparisonTableWithHC}. For each graph, we list the known chromatic number $\chi$ or the best $k^*$ reported in the literature when $\chi$ is still unknown. For each algorithm, we indicate the number of the best (the smallest) $k$ value for which the algorithm attains a legal $k$-coloring and the number of such successful runs over 20 executions (\#hit). The differences between the best $k$ of $\rm{RLS}_{\rm{0}}$ and the best $k$ of RLS are provided in the last column. The results show that RLS significantly outperforms $\rm{RLS}_{\rm{0}}$ in terms of the best $k$ value for 29 out of 32 instances (indicated in bold). For example, on instance \texttt{flat\_26\_0}, RLS attains the chromatic number $k$ (i.e., $\chi = 26$) while $\rm{RLS}_{\rm{0}}$ needs 45 colors to color it legally. Specially, we observe that RLS has a larger improvement on hard instances than on easy instances. For instance, \texttt{latin\_square\_10} and \texttt{flat\_76\_0} are two well-known hard instances, RLS achieves two largest improvements, i.e., using 70 and 46 fewer colors than $\rm{RLS}_{\rm{0}}$. In summary, RLS attains better results on 29 out of 32 instances compared to its variant with the reinforcement learning mechanism disabled. This experiment confirms the effectiveness of the reinforcement learning mechanism to help the descent-based local search to attain much better results.

\subsubsection{Effectiveness of the probability smoothing technique}
\label{Effectiveness of the probability smoothing technique}

To study the effectiveness of the probability smoothing technique used in RLS, we compare RLS with its alternative algorithm $\rm{RLS}_{\rm{1}}$, which is obtained from RLS by adjusting the probability updating scheme. More specifically, $\rm{RLS}_{\rm{1}}$ works in the same way as RLS, but it does not use the probability smoothing strategy, that is, line 12 in Algorithm \ref{algorithm:RLS} is removed. For this experiment, by following \cite{Galinier1999}, we use \textit{running profiles} to observe the change of evaluation function $f$ over the number of iterations. Running profiles provide interesting information about the convergence of the studied algorithms. 

\begin{figure}[!htbp]
\centering
\includegraphics[width=1.0\textwidth]{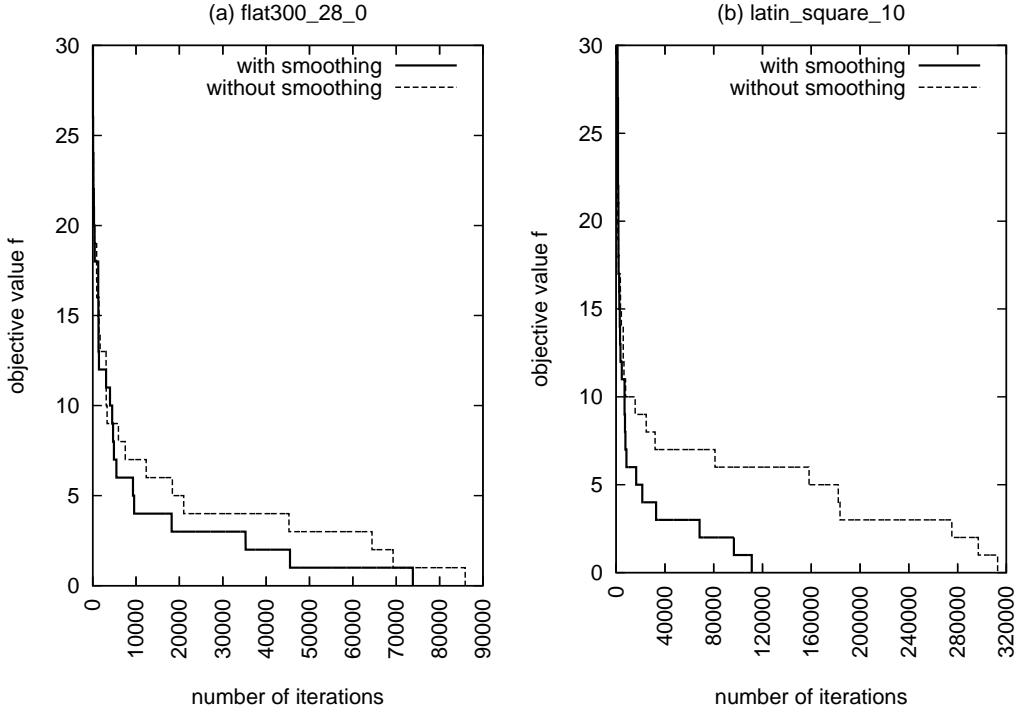}
\caption{Running profile of RLS (with smoothing) and $\rm{RLS}_{\rm{1}}$ (without smoothing) on instance \texttt{flat300\_28\_0} and \texttt{latin\_square\_10}}\label{fig:smoothing}
\end{figure}

The running profiles of RLS and $\rm{RLS}_{\rm{1}}$ are shown in Figure \ref{fig:smoothing} on two selected instances: Fig. \ref{fig:smoothing}(a) for \texttt{flat300\_28\_0} ($k = 32$), and Fig.\ref{fig:smoothing}(b) for \texttt{latin\_square\_10} ($k = 101$). We observe that though both algorithms successfully obtain a legal $k$-coloring, RLS converges to the best solution more quickly than $\rm{RLS}_{\rm{1}}$, i.e., the objective value $f$ of RLS decreases more quickly than that of $\rm{RLS}_{\rm{1}}$. Consequently, RLS needs less iterations to attain a legal solution. This experiment demonstrated the benefit of using probability smoothing technique in RLS. 


\subsubsection{Comparison of different group selection strategies}
\label{Comparison of different group selection strategies}

The group selection strategy plays an important role in RLS. At each iteration, each vertex selects a suitable group based on the group selection strategy to produce a new solution for the next round of the descent-based local search optimization. In this section, we show an analysis of the group selection strategies to confirm the interest of the adopted hybrid strategy which combines random and greedy strategies. 

The investigation was carried out between RLS and its variant $\rm{RLS}_{\rm{2}}$, which is obtained from RLS by means of replacing the hybrid group selection strategy with the roulette wheel selection strategy. In the experiment, each instance was tested 20 times independently with different random seeds. The number of successful runs, the average number of iterations and the average running time of successful runs are reported. 

\begin{table}[!htbp]
\caption{Comparative performance of RLS (with its hybrid group selection strategy) and $\rm{RLS}_{\rm{2}}$ (with the roulette wheel selection strategy). Smaller $k$ and larger \#hit are better}
\label{tab:aComparisonTableWithGSS}
\begin{center}
\begin{scriptsize}
\begin{tabular}{lrrrrrrr}
\toprule[0.75pt]
\multicolumn{1}{c}{} & \multicolumn{3}{c}{$\rm{RLS}_{\rm{2}}$} && \multicolumn{3}{c}{RLS}\\
\cmidrule[0.5pt]{2-4} \cmidrule[0.5pt]{6-8}
Instance   & $k_1$(\#hit)& \#iter 		  & time(s) && $k_2$(\#hit) & \#iter         & time(s)\\
\midrule[0.5pt]
le450\_25c & 26(0/20)  & -                 & -       && 26(13/20) & $4.7 \times 10^6$ & 181.39 \\
 		   & 27(20/20) & $7.0 \times 10^5$ & 26.86   && 27(20/20) & $1.5 \times 10^6$ & 61.13  \\
DSJR500.1  & 12(0/20)  & -                 & -       && 12(20/20) & $7.8 \times 10^4$ & 1.91	 \\
           & 13(20/20) & $2.0 \times 10^6$ & 50.42   && 13(20/20) & $3.0 \times 10^3$ & 0.10   \\
DSJR500.1c & 85(0/20)  & -                 & -       && 85(02/20) & $4.6 \times 10^6$ & 699.63 \\
           & 86(0/20)  & -                 & -       && 86(20/20) & $3.6 \times 10^6$ & 529.47 \\
           & 87(20/20) & $3.2 \times 10^6$ & 361.97  && 87(20/20) & $6.9 \times 10^5$ & 108.12 \\
DSJC1000.1 & 21(09/20) & $2.0 \times 10^7$ & 1508.48 && 21(20/20) & $1.4 \times 10^7$ & 1223.18\\
           & 22(20/20) & $6.0 \times 10^5$ & 41.82   && 22(20/20) & $8.0 \times 10^5$ & 64.77  \\
\bottomrule[0.75pt]
\end{tabular}
\end{scriptsize}
\end{center}
\end{table} 

Table \ref{tab:aComparisonTableWithGSS} show comparative results of RLS with $\rm{RLS}_{\rm{2}}$ for the chosen instances. The results indicate that RLS significantly outperforms $\rm{RLS}_{\rm{2}}$ in terms of the best $k$ value and the number of successful running times. For example, on instance \texttt{DSJR500.1c}, RLS colors this graph with 85 colors, while $\rm{RLS}_{\rm{2}}$ needs more colors ($k = 87$) to color it. A similar observation can be found on instance \texttt{le450\_25c}, for which RLS obtains a legal $26$-coloring, while $\rm{RLS}_{\rm{2}}$ only obtains a $27$-coloring. Furthermore, when they need the same number of colors to color a graph \texttt{DSJC1000.1}, RLS achieves it with a higher success rate compared to $\rm{RLS}_{\rm{2}}$. This experiment confirms the interest of the adopted hybrid selection strategy.

\subsection{Computational results of RLS and comparisons}
\label{Comparisons with other local search algorithms}

We turn now our attention to a comparative study of the proposed RLS approach with respect to some well-known coloring algorithms in the literature. This study focuses on five algorithms based on advanced local search methods including the prominent simulating annealing (SA) algorithm \cite{Johnson1991}, the improved tabu search (TS) algorithm \cite{Galinier1999}, the guided local search (GLS) algorithm \cite{Chiarandini2005}, the iterative local search (ILS) algorithm \cite{Chiarandini2002} and the focused walk based local search (FWLS) algorithm \cite{Wu2013}. This comparison is not exhaustive, yet it allows us to assess the effectiveness of using the learning mechanism to boost a very simple descent procedure.  


\begin{table}[!htbp]
\caption{Comparative results of RLS and five local search algorithms on DIMACS graphs}
\label{tab:aSummaryTable1}
\begin{center}
\begin{scriptsize}
\begin{tabular}{lrrrrrrrrr}
\toprule[0.75pt]
\multirow{2}{*}{Instance} & \multirow{2}{*}{$n$} & \multirow{2}{*}{$m$} & \multirow{2}{*}{$\chi/k^*$} & RLS & TS \cite{Galinier1999} & SA \cite{Johnson1991} & GLS \cite{Chiarandini2005} & ILS \cite{Chiarandini2002} & FWLS \cite{Wu2013}\\
\cline{5-10}
& & & & & 1999 & 1991 &  2005 & 2002 & 2013\\
\midrule[0.5pt]
DSJC125.1  		& 125 & 736     & ?/5	& \textbf{5}  & 5  & 6  & 5   & 5   & 5  \\
DSJC125.5  		& 125 & 3,891   & ?/17	& \textbf{17} & 17 & 18 & 18  & 17  & 17 \\
DSJC125.9  		& 125 & 6,961   & ?/44	& \textbf{44} & 44 & 44 & 44  & 44  & 45 \\
DSJC250.1  		& 250 & 3,218   & ?/8	& \textbf{8}  & 8  & 9  & 8   & 8   & 8  \\
DSJC250.5  		& 250 & 15,668  & ?/28	& 29          & 29 & 29 & 29  & 28  & 29 \\
DSJC250.9  		& 250 & 27,897  & ?/72	& 75          & 72 & 72 & 72  & 72  & 73 \\
DSJC500.1  		& 500 & 12,458  & ?/12	& \textbf{13} & 13 & 14 & 13  & 13  & 13 \\
DSJC500.5  		& 500 & 62,624  & ?/47	& \textbf{50} & 50 & 51 & 52  & 50  & 51 \\
DSJC1000.1 		& 1,000 & 49,629 & ?/20	& \textbf{21} & 21 & 23 & 21  & 21  & 21 \\ 
DSJR500.1       & 500 & 3,555   & ?/12	& \textbf{12} & 12 & -  & -   & 12  & -  \\
DSJR500.1c      & 500 & 121,275 & ?/84	& \textbf{85} & 94 & 89 & 85  & -   & -  \\
flat300\_20\_0  & 300 & 21,375 & 20/20	& \textbf{20} & 20 & 20 & 20  & 20  & -  \\
flat300\_26\_0  & 300 & 21,633 & 26/26	& \textbf{26} & 26 & 32 & 33  & 26  & 26 \\
flat300\_28\_0  & 300 & 21,695 & 28/28	& 32          & 32 & 33 & 33  & 31  & 28 \\
flat1000\_76\_0 & 1,000 & 246,708 & ?/81	& \textbf{89} & 91 & 89 & 92  & 89  & 90 \\
le450\_15a 		& 450 & 8,168  & 15/15	& \textbf{15} & 15 & 16 & 15  & 15  & 15 \\
le450\_15b 		& 450 & 8,169  & 15/15	& \textbf{15} & 15 & 16 & 15  & 15  & 15 \\
le450\_15c 		& 450 & 16,680 & 15/15	& \textbf{15} & 16 & 23 & 15  & 15  & 15 \\
le450\_15d 		& 450 & 16,750 & 15/15	& \textbf{15} & 16 & 22 & 15  & 15  & 15 \\
le450\_25a 		& 450 & 8,260  & 25/25	& 26          & 25 & -  & -   & -   & 25 \\
le450\_25b 		& 450 & 8,263  & 25/25	& \textbf{25} & 25 & -  & -   & -   & 25 \\
le450\_25c 		& 450 & 17,343 & 25/25	& \textbf{26} & 26 & 27 & 26  & 26  & 26 \\
le450\_25d 		& 450 & 17,425 & 25/25	& \textbf{26} & 26 & 28 & 26  & 26  & 26 \\
R125.1  			& 125 & 209    & ?/5		& \textbf{5}  & 5  & -  & -   & -   & -  \\
R125.1c 			& 125 & 7,501  & ?/46	& \textbf{46} & 47 & -  & -   & -   & -  \\
R125.5  			& 125 & 3,838  & ?/36	& 38          & 36 & -  & -   & -   & -  \\
R250.1  			& 250 & 867    & ?/8		& \textbf{8}  & 8  & -  & -   & -   & -  \\
R250.1c 			& 250 & 30,227 & ?/64	& \textbf{64} & 72 & -  & -   & -   & -  \\
R1000.1 			& 1,000 & 14,348 & ?/20	& 21 		  & 20 & -  & -   & -   & -  \\
school1         & 385 & 19,095 & ?/14	& \textbf{14} & 14 & 14 & 14  & -   & 14 \\
school1\_nsh		& 352 & 14,612 & ?/14	& \textbf{14} & 14 & 14 & 14  & -   & 14 \\
latin\_square\_10 & 900 & 307,350 & ?/97	& \textbf{99} & 105&101 & 102 & 103 & -  \\
\midrule[0.5pt]
\multicolumn{3}{c}{better} & 0/32		& -  		  & 7/32   & 16/23 & 5/22  & 1/21  & 3/22  \\
\multicolumn{3}{c}{equal}  & 18/32		& -			  & 21/32  & 6/23  & 16/22 & 17/21 & 17/22 \\
\multicolumn{3}{c}{worse}  & 14/32		& -			  & 4/32   & 1/23  & 1/22  & 3/21  & 2/22  \\
\bottomrule[0.75pt]
\end{tabular}
\end{scriptsize}
\end{center}
\end{table}
We present in Tables \ref{tab:aSummaryTable1} and \ref{tab:aSummaryTable2} the results of RLS together with the best solutions of these algorithms. We list the number of vertices ($n$) and edges ($m$) of each graph, the known chromatic number $\chi$ or the best $k^*$ reported in the literature when $\chi$ is still unknown. For each algorithm, we list the best (the smallest) $k$ for which a legal $k$-coloring is attained. A summary of the comparisons between our RLS algorithm and each reference algorithm is provided at the bottom of these tables. The rows `better', `equal', and `worse' respectively represent the number of instances for which our RLS algorithm achieves a better, an equal, and a worse solution than the corresponding reference algorithm over the total number of instances for which the algorithm is tested. The results of the reference algorithms are extracted from the literature except for TS which was run on the same computing platform as RLS. In these tables, `$-$' indicate that the result of the algorithm on this instance is unavailable in the literature. When a result of RLS is no worse than any result of the competing algorithms, this result is marked in bold.

From Table \ref{tab:aSummaryTable1} which concerns the DIMACS graphs, we observe that our RLS algorithm achieves a competitive and even better performance on some graphs. For instance, compared to SA, RLS finds 16 better best solutions out of the $23$ instances tested by SA. The comparison with TS is more informative and meaningful given that RLS and TS share the same data structures, both were programmed in C and were run on the same computer. We observe that despite its simple descent local search procedure, RLS attains better results than TS thanks to its learning mechanism. Additionally, we observe that RLS achieves much more `better' results than `worse' results compared to the reference algorithms except for ILS. Nevertheless, since the results of ILS for 9 instances are unavailable, it is difficult to draw a clear conclusion.

\begin{table}[!htbp]
\caption{Comparative results of RLS and other algorithms on COLOR02 graphs}
\label{tab:aSummaryTable2}
\begin{center}
\begin{scriptsize}
\begin{tabular}{lrrrrrrrrr}
\toprule[0.75pt]
\multirow{2}{*}{Instance} & \multirow{2}{*}{$n$} & \multirow{2}{*}{$m$} & \multirow{2}{*}{$\chi/k^*$} & RLS & TS \cite{Galinier1999} & SA \cite{Johnson1991} & GLS \cite{Chiarandini2005} & ILS \cite{Chiarandini2002} & FWLS \cite{Wu2013}\\
\cline{5-10}
& & & & & 1999 & 1991 & 2005 & 2002 & 2013\\
\midrule[0.5pt]
miles250 & 128 & 387 & 8/8 			   & \textbf{8}(20/20)	& 8  & -  & -  & -  & -  \\
miles500 & 128 & 1,170 & 20/20 		   & \textbf{20}(20/20)	& 20 & -  & -  & -  & -  \\
miles750 & 128 & 2,113 & 31/31 	       & \textbf{31}(20/20)	& 31 & -  & -  & -  & -  \\
miles1000 & 128 & 3,216 & 42/42  	   & \textbf{42}(20/20)	& 42 & -  & -  & -  & -  \\
miles1500 & 128 & 5,189 & 73/73 	   	   & \textbf{73}(20/20)	& 73 & -  & -  & -  & -  \\
myciel3   & 11  & 20    & 4/4 		   & \textbf{4}(20/20)	& 4  & -  & -  & -  & 4  \\
myciel4 & 23 & 71 & 5/5 			       & \textbf{5}(20/20)	& 5  & -  & -  & -  & 5  \\
myciel5 & 47 & 236 & 6/6 			   & \textbf{6}(20/20)	& 6  & -  & -  & -  & 6  \\
myciel6 & 95 & 755 & 7/7 			   & \textbf{7}(20/20)	& 7  & -  & -  & -  & 7  \\
myciel7 & 191 & 2,360 & 8/8 		       & \textbf{8}(20/20)	& 8  & -  & -  & -  & 8  \\
queen5\_5 & 25 & 160 & ?/5 			   & \textbf{5}(20/20)	& 5  & -  & -  & -  & 5  \\
queen6\_6 & 36 & 290 & ?/7 			   & \textbf{7}(20/20)	& 7  & 7  & 7  & 7  & 7  \\
queen7\_7 & 49 & 476 & ?/7 			   & \textbf{7}(20/20)	& 7  & 7  & 7  & 7  & 7  \\
queen8\_8 & 64 & 728 & ?/9 			   & \textbf{9}(20/20)	& 9  & 9  & 9  & 9  & 9  \\
queen8\_12 & 96 & 1,368 & ?/12 		   & \textbf{12}(20/20)	& 12 & 12 & 12 & 12 & 12 \\
queen9\_9 & 81 & 2,112 & ?/10 		   & \textbf{10}(20/20)	& 10 & 10 & 10 & 10 & 10 \\
queen10\_10 & 100 & 2,940 & ?/11 	   & \textbf{11}(20/20)	& 11 & 11 & 11 & 11 & 11 \\
queen11\_11 & 121 & 3,960 & ?/11 	   & \textbf{12}(20/20) 	& 12 & 12 & 12 & 12 & 12 \\
queen12\_12 & 144 & 5,192 & ?/12 	   & \textbf{13}(20/20)	& 13 & 14 & 13 & 13 & 13 \\
queen13\_13 & 169 & 6,656 & ?/13 	   & \textbf{14}(20/20)	& 14 & 15 & 15 & 14 & 14 \\
queen14\_14 & 196 & 8,372 & ?/14 	   & \textbf{15}(14/20)	& 15 & 16 & 16 & 15 & 15 \\
queen15\_15 & 225 & 10,360 & ?/15 	   & \textbf{16}(11/20)	& 16 & 17 & 17 & 16 & 16 \\
queen16\_16 & 256 & 12,640 & ?/16 	   & \textbf{17}(11/20)	& 18 & 17 & 18 & 18 & 18 \\
\midrule[0.5pt]
\multicolumn{3}{c}{better}& 0/23  	   & -					& 1/23  & 4/12 & 4/12 & 1/12  & 1/18  \\
\multicolumn{3}{c}{equal} & 17/23	   & -					& 22/23 & 8/12 & 8/12 & 11/12 & 17/18 \\
\multicolumn{3}{c}{worse} & 6/23		   & -					& 0/23  & 0/12 & 0/12 & 0/12  & 0/18  \\
\bottomrule[0.75pt]
\end{tabular}
\end{scriptsize}
\end{center}
\end{table}

When we compare our RLS algorithm with the five reference algorithms on the COLOR02 graph instances (Table \ref{tab:aSummaryTable2}), we observe that RLS dominates these algorithms. Specifically, RLS achieves no worse results on these instances than any of the reference algorithms, and obtains $4$ better solutions than SA and GLS, $1$ better solution than TS, ILS, and FWLS respectively.  

We also find that our proposed RLS method even achieves competitive performances compared to some complex population algorithms proposed in recent years, such as ant-based algorithm \cite{Bui2008} (2008), and modified cuckoo optimization algorithm \cite{Mahmoudi2015} (2015). However, given the very simplicity of its underlying local search procedure, it is no surprise that RLS alone cannot compete with the most powerful coloring algorithms like \cite{Galinier1999,Lu2010,Malagutietal2008,Porumbel2010b,Titiloye2011,WuHao2012}. Indeed, these algorithm are typically complex hybrid algorithms mixing several approaches like genetic computing and local optimization. On the other hand, given the way the proposed RLS approach is composed, it would be interesting to replace the simple descent-based local search by any of these advanced coloring algorithms and investigate the proposed reinforcement learning mechanism in comparison with these advanced coloring algorithms.

\section{Conclusion and discussion}
\label{sec:Conclusion}

In this paper, we proposed a reinforcement learning based optimization approach for solving the class of grouping problems. The proposed RLS approach combines reinforcement learning techniques with a descent-based local search procedure. Reinforcement learning is used to maintain and update a set of probability vectors, each probability vector specifying the probability that an item belongs to a particular group. At each iteration, RLS builds a starting grouping solution according to the probability vectors and with the help of a group selection strategy. RLS then applies a descent-based local search procedure to improve the given grouping solution until a local optimum is reached. At this point, the starting solution and the ending local optimum solution are compared to update the probability vector of each item according to the situation of the item. Specifically, RLS rewards the selected group of the item if the item stays in the original group, otherwise RLS penalizes the selected group and compensates the new group.


Experimental analyses and performance assessments of the RLS approach were carried out on the graph coloring problem which is a well-known grouping problem. Based on experimental results on  popular DIMACS and COLOR02 benchmark graphs, we showed that 1) reinforcement learning is highly valuable to increase the performance of the descent-based local search procedure; 2) the probability smoothing technique which forgets old decisions is very useful to avoid search traps; and 3) the hybrid group selection strategy combining randomness and greediness is more suitable than other selection strategies. 

In terms of computational results, RLS, despite the simplicity of its basic coloring procedure, proved to be competitive compared to five advanced local search algorithms. It performs even better than some recent and complex optimization algorithms like the ant-based algorithm \cite{Bui2008} and modified cuckoo optimization algorithm \cite{Mahmoudi2015}. On the other hand, given the competitiveness of the graph coloring problems, RLS cannot really competes with the most advanced coloring algorithms which are often based on complex hybrid schemes. Fortunately, given the way of reinforcement learning being used in RLS, it is reasonable to believe that the proposed reinforcement learning techniques could be combined with these advanced coloring approaches, e.g., by replacing the descent procedure with a more powerful algorithm within the RLS approach. Such a possibility constitutes one of our future research. 

Finally, another future work is to apply the proposed approach to solve other grouping problems. For this purpose, it is necessary to devise a descent local search procedure (or any other solution improvement procedure) for the studied problem while the other ingredients of the RLS approach can be kept unchanged.

\section*{Acknowledgments}
This work was partially supported by the PGMO project (2014-2015, Jacques Hadamard Mathematical Foundation, Paris). The financial support for Yangming Zhou from the China Scholarship Council (CSC, 2014-2018) is acknowledged.


\begin{thebibliography}{1}
\bibitem{Agust2012}
Agustı´n-Blas, L., Salcedo-Sanz, S., Jim{\'e}nez-Fern{\'a}ndez, S., Carro-Calvo, L., Ser, J. D. \& Portilla-Figueras, J. (2012), A new grouping genetic algorithm for clustering problems, Expert Systems with Applications, 39, 9695-9703.

\bibitem{AkbariTorkestani2011} 
Torkestani, J. A. \& Meybodi, M. R. (2011), A cellular learning automata-based algorithm for solving the vertex coloring problem, Expert Systems with Applications, 38, 9237-9247.

\bibitem{Baluja2000}
Baluja, S., Barto, A., Boese, K., Boyan, J., Buntine, W., Carson, T., Caruana, R., Cook, D., Davies, S., Dean, T., \& others (2000). Statistical machine learning for large-scale optimization, Neural Computing Surveys, 3, 1-58.

\bibitem{Battiti2014}
Battiti, R., \& Brunato, M. (2014). The LION way. Machine Learning plus Intelligent Optimization, LIONlab, University of Trento.

\bibitem{Boyan2001}
Boyan, J.A., \& Moore, A. W. (2001). Learning evaluation functions to improve optimization by local search, The Journal of Machine Learning Research, 1, 77-112.

\bibitem{Bui2008}
Bui, T.N., Nguyen, T.H., Patel, C.M., \& Phan, K. T. (2008). An ant-based algorithm for coloring graphs, Discrete Applied Mathematics, 156, 190-200.

\bibitem{Burke2003}
Burke, E., Kendall, G., \& Soubeiga, E. (2003). A Tabu-Search Hyper-heuristic for Timetabling and Rostering, Journal of Heuristics, 9, 451-470.


\bibitem{Cai2013}
Cai, S., Su, K., Luo, C., \& Sattar, A. (2013). NuMVC: An efficient local search algorithm for minimum vertex cover, Journal of Artificial Intelligence Research, 687-716.

\bibitem{Ceberio2013}
Ceberio, J., Mendiburu, A., \& Lozano, J. A. (2013). The Plackett-Luce ranking model on permutation-based optimization problems, Proceedings of the IEEE Congress on Evolutionary Computation (CEC), 494-501.

\bibitem{Chiarandini2005}
Chiarandini, M. (2005). Stochastic local search methods for highly constrained combinatorial optimisation problems, Ph.D. thesis, Technical University of Darmstadt.

\bibitem{Chiarandini2002}
Chiarandini, M., \& St\"{u}tzle, T. (2002). An application of iterated local search to graph coloring problem, Proceedings of the Computational Symposium on Graph Coloring and its Generalizations, 112-125.

\bibitem{Elhag2O15}
Elhag, A., \& \"{O}zcan, E. (2015). A grouping hyper-heuristic framework: Application on graph coloring, Expert Systems with Applications, 42, 5491-5507.

\bibitem{Falkenauer1998}
Falkenauer, E. (1998). Genetic Algorithms and Grouping Problems, John Wiley \& Sons, Inc.

\bibitem{Fleurent1996}
Fleurent, C., \& Ferland, J. (1996). Genetic and hybrid algorithms for graph coloring, Annals of Operations Research, 63, 437-461.

\bibitem{Galinier1999}
Galinier, P., \& Hao, J.K. (1999). Hybrid evolutionary algorithms for graph coloring, Journal of Combinatorial Optimization, 3, 379-397.

\bibitem{Galinier2013}
Galinier, P., Hamiez, J.P., Hao, J.K., \& Porumbel, D. (2013). Recent advances in graph vertex coloring, In Handbook of optimization, 505-528.

\bibitem{GareyJohnson1979}
Garey,  M., \& Johnson,  D. (1979).  Computers  and  Intractability:  A  Guide  to  the  Theory  of  NP-Completness. W.H. Freeman and Co., San Franc., USA.

\bibitem{Gosavi2009}
Gosavi, A. (2009). Reinforcement learning: A tutorial survey and recent advances. INFORMS Journal on Computing, 21, 178-192.

\bibitem{Guoetal2013}
Guo, Y., Goncalves, G., \& Hsu, T. (2013). A multi-agent based self-adaptive genetic algorithm for the long-term car pooling problem, Journal of Mathematical Modelling and Algorithms in Operations Research, 12(1), 45-66.

\bibitem{Hafiz2016}
Hafiz, F. \& Abdennour, A. (2016), Particle Swarm Algorithm variants for the Quadratic Assignment Problems - A probabilistic learning approach, Expert Systems with Applications, 2016, 44, 413-431.

\bibitem{HamiezHao1993}
Hamiez, J.P., \& Hao, J.K. (1993), An analysis of solution properties of the graph coloring problem. Metaheuristics: Computer Decision-Making, Chapter 15, pp325-346, Resende M.G.C. and de Sousa J.P. (Eds.), Kluwer.

\bibitem{Hertz1987}
Hertz, A., \& de Werra, D. (1987). Using tabu search techniques for graph coloring, Computing, 39, 345-351.

\bibitem{Hutter2002}
Hutter, F., Tompkins, D.A.D., \& Hoos, H. H. (2002). Scaling and probabilistic smoothing: efficient dynamic local search for SAT, Principles and Practice of Constraint Programming (CP), pp. 233-248.

\bibitem{Hutter2014}
Hutter, F., Xu, L., Hoos, H.H., \& Leyton-Brown, K. (2014). Algorithm runtime prediction: methods \& evaluation, Artificial Intelligence, 206, 79-111.

\bibitem{Ishtaiwi2005}
Ishtaiwi, A., Thornton, J., Sattar, A., \& Pham, D. N. (2005). Neighbourhood clause weight redistribution in local search for SAT, Principles and Practice of Constraint Programming (CP), pp. 772-776.

\bibitem{Johnson1991}
Johnson, D.S., Aragon, C.R., McGeoch, L.A., \& Schevon, C. (1991). Optimization by simulated annealing: an experimental evaluation; part II, graph coloring and number partitioning, Operations Research, 39, 378-406.

\bibitem{JohnsonTrick1996}
Johnson, D.S., \&  Trick  M. (1996). Cliques, Coloring, and Satisfiability, DIMACS Series in Discrete Math. and Theor. Comput. Sci. , vol. 26, pp. 335-357. Am. Math. Soc., New Providence, USA.


\bibitem{Kashan2013}
Kashan, A. H., Kashan, M.H., \& Karimiyan, S. (2013). A particle swarm optimizer for grouping problems, Information Sciences, 252, 81-95.


\bibitem{Lewis2009}
Lewis, R. (2009). A general-purpose hill-climbing method for order independent minimum grouping problems: A case study in graph colouring and bin packing, Computers \& Operations Research, 36(7), 2295-2310.

\bibitem{Lewis2007}
Lewis, R., \& Paechter, B. (2007). Finding feasible timetables using group-based operators, Evolutionary Computation, IEEE Transactions on, 11(3), 397-413.


\bibitem{Lu2010}
L\"{u}, Z., \& Hao, J. (2010). A memetic algorithm for graph coloring, European Journal of Operational Research, 203, 241-250.

\bibitem{Malagutietal2008}
Malaguti, E., Monaci, M., \& Toth, P. (2008). A metaheuristic approach for the vertex coloring problem, INFORMS Journal on Computing, 20(2), 302-316.

\bibitem{MalagutiToth2009}
Malaguti, E., \& Toth, P. (2009). A survey on vertex coloring problems, International Transactions in Operational Research, 17(1), 1-34.

\bibitem{Mahmoudi2015}
Mahmoudi, S., \& Lotfi, S. (2015). Modified cuckoo optimization algorithm (MCOA) to solve graph coloring problem, Applied Soft Computing, 33, 48-64.

\bibitem{Miagkikh1999}
Miagkikh, V.V., \& Punch III, W. F. (1999). Global search in combinatorial optimization using reinforcement learning algorithms, Proceedings of the Congress on Evolutionary Computation (CEC).

\bibitem{Porumbel2010a}
Porumbel, D.C., Hao, J.K., \& Kuntz, P. (2010a). A search space ``cartography'' for guiding graph coloring heuristics, Computers \& Operations Research, 37, 769-778.

\bibitem{Porumbel2010b}
Porumbel, D.C., Hao, J.-K., \& Kuntz, P. (2010b). An Evolutionary Approach with Diversity Guarantee and Well-Informed Grouping Recombination for Graph Coloring, Computers \& Operations Research 37(10), 1822--1832.

\bibitem{Quiroz-Castellanos2015}
Quiroz-Castellanos, M., Cruz-Reyes, L., Torres-Jim{\'e}nez, J., G{\'o}mez, C., Huacuja, H.J., \& Alvim, A.C. F. (2015). A grouping genetic algorithm with controlled gene transmission for the bin packing problem, Computers \& Operations Research, 55, 52-64.

    

\bibitem{Schuurams2001}
Schuurmans, D.,  Southey, F., \& Holte, R.C.  (2001). The exponentiated subgradient algorithm for heuristic boolean programming, In Proceedings of the 7th International Joint Conference on Artificial Intelligence (IJCAI), pp. 334-341.

\bibitem{Sghiretal2015}
Sghir, I., Hao, J.K., Jaafar, I.B., \&  Ghédira, K. (2015). A multi-agent based optimization method applied to the quadratic assignment problem, Expert Systems with Applications, 42(23), 9252-9263.

\bibitem{Thornton2004}
Thornton, J., Duc, N.P., Stuart B., \&   Valnir F.Jr. (2004). Additive versus multiplicative clause weighting for SAT, In Proceedings of the Conference of the American Association for Artificial Intelligence (AAAI), pp. 191-196.

\bibitem{Titiloye2011}
Titiloye, O., \& Crispin, A. (2011). Quantum annealing of the graph coloring problem, Discrete Optimization, 8, 376-384.


\bibitem{WuHao2012}
Wu, Q., \& Hao, J.K. (2012). Coloring large graphs based on independent set extraction, Computers  \& Operations Research, 39(2), 283-290.

\bibitem{Wu2013}
Wu, W., Luo, C., \& Su, K. (2013). FWLS: A Local Search for Graph Coloring, Frontiers in Algorithmics and Algorithmic Aspects in Information and Management, Third Joint International Conference, Proceedings, pp.84-93.


\bibitem{Xu2009}
Xu, Y., Stern, D., \& Samulowitz, H. (2009), Learning adaptation to solve constraint satisfaction problems, Proceedings of Learning and Intelligent Optimization (LION III), Jan 14-18, Trento, Italy.

\end{thebibliography}
\end{document}